\documentclass{article}

\usepackage[final]{corl_2020} %
\usepackage{url}            %
\usepackage{booktabs}       %
\usepackage{amsfonts}       %
\usepackage{amsmath,amssymb}
\usepackage{gensymb}
\usepackage{graphicx}
\usepackage[noabbrev,capitalise]{cleveref}
\usepackage{nicefrac}       %
\usepackage{microtype}      %
\usepackage[export]{adjustbox}
\usepackage{wrapfig}

\newcommand{\algname}{\mbox{MATS}}

\title{\algname{}: An Interpretable Trajectory Forecasting Representation for Planning and Control}

\author{
  Boris Ivanovic$^1$ \hspace{0.3cm} Amine Elhafsi$^1$ \hspace{0.3cm} Guy Rosman$^2$ \hspace{0.3cm} Adrien Gaidon$^2$ \hspace{0.3cm} Marco Pavone$^1$  \\
  $^1$Stanford University \hspace{1cm} $^2$Toyota Research Institute\\
  \texttt{\{borisi, amine, pavone\}@stanford.edu} \hspace{0.2cm} \texttt{\{guy.rosman, adrien.gaidon\}@tri.global}\\
}

\begin{document}
\maketitle

\begin{abstract}
    Reasoning about human motion is a core component of modern human-robot interactive systems. 
    In particular, one of the main uses of behavior prediction in autonomous systems is to inform robot motion planning and control. 
    However, a majority of planning and control algorithms reason about system dynamics rather than the predicted agent tracklets (i.e., ordered sets of waypoints) that are commonly output by trajectory forecasting methods, which can hinder their integration.
    Towards this end, we propose Mixtures of Affine Time-varying Systems (MATS) as an output representation for trajectory forecasting that is more amenable to downstream planning and control use. 
    Our approach leverages successful ideas from probabilistic trajectory forecasting works to learn dynamical system representations that are well-studied in the planning and control literature. 
    We integrate our predictions with a proposed multimodal planning methodology and demonstrate significant computational efficiency improvements on a large-scale autonomous driving dataset.
\end{abstract}


\section{Introduction}\label{sec:introduction}
Reasoning about human motion is an important prerequisite to safe and socially-aware robot navigation. As a result, multi-agent behavior prediction has become a core component of modern human-robot interactive systems such as self-driving cars.
In particular, one of the main uses of behavior prediction in autonomous systems is to inform robot motion planning and control.
Accordingly, there have been many approaches tackling human motion prediction specifically. Earlier models were predominantly deterministic regressors leveraging state-space models~\cite{HelbingMolnar1995}, Gaussian process regression~\cite{WangFleetEtAl2008}, inverse reinforcement learning~\cite{LeeKitani2016}, game theory~\cite{LiYaoEtAl2019}, or recurrent neural networks (RNNs) \cite{AlahiGoelEtAl2016}. Recent works focus on capturing the multimodality of human behavior, developing deep generative models that leverage conditional variational autoencoders (CVAEs) \cite{SohnLeeEtAl2015} to explicitly capture multimodality \cite{LeeChoiEtAl2017,IvanovicSchmerlingEtAl2018,DeoTrivedi2018,SadeghianLegrosEtAl2018,IvanovicPavone2019,SalzmannIvanovicEtAl2020}, or generative adversarial networks \cite{GoodfellowPouget-AbadieEtAl2014} to implicitly do so \cite{GuptaJohnsonEtAl2018,ZhaoXuEtAl2019}. A unifying theme among these otherwise disparate works is that they produce trajectories (or distributions thereof) for each agent in a scene, an intuitive output representation that matches common evaluation metrics. However, a majority of planning and control algorithms reason about \textit{system dynamics} rather than future agent tracklets (i.e., ordered sets of waypoints), which can hinder the integration of state-of-the-art trajectory forecasting methods in planning and control.
For example, sampling future trajectory distributions to obtain rare events may be too expensive in a fixed-computation, safety-critical setting such as autonomous driving. 

\textbf{Contributions.} To bridge this gap, we propose \textit{Mixtures of Affine Time-varying Systems (\algname{})}\footnote{All of our code, models, and data can be found online at \url{https://github.com/StanfordASL/MATS}.} as an output representation for trajectory forecasting that 
is significantly more amenable to downstream planning and control use. Our contributions are two-fold. First, we show that \algname{}, a linear-affine dynamical structure, is a viable trajectory forecasting representation even for highly non-linear real-world systems. Second, we show that such a prediction representation yields significant reductions in downstream planning and control complexity, even when accounting for multimodal predictions (i.e., the potential for many possible futures), at the cost of a minor regression in raw prediction performance due to the added structure.

\begin{figure}[t]
    \centering
    \includegraphics[width=\linewidth]{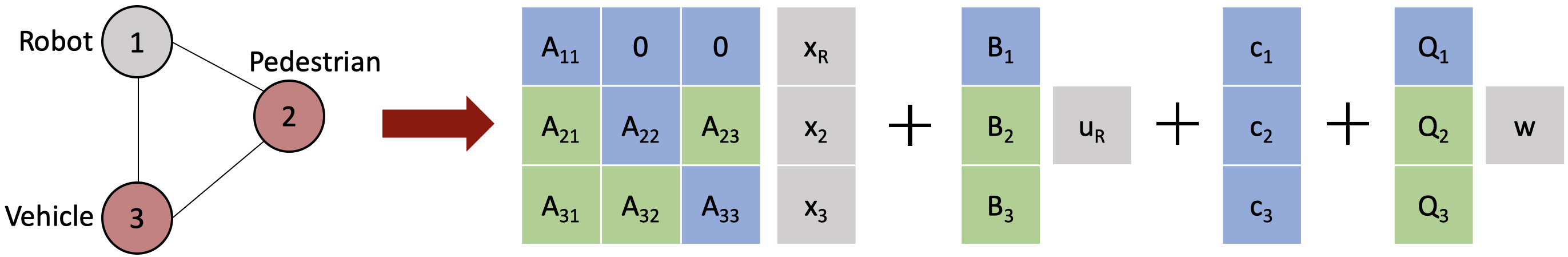}
    \caption{Our model represents edges in a spatiotemporal graph as blocks of dynamical system matrices. Blocks that are fully determined by dynamics (e.g., $A$'s diagonal and $c$) or solely involve the ego-robot (e.g., the first block of $B$) are not learned (blue). All other blocks model agent-agent interactions or uncertainty, and are learned (green). Note that off-diagonal blocks in the ego-robot's row in $A$ are zero, encoding the fact that the ego-robot is only controlled by its actions, $\mathbf{u}_\text{R}$.}
    \label{fig:architecture}
\end{figure}

\section{Related Work}\label{sec:related_work}

\textbf{Integrating Trajectory Forecasting with Motion Planning.} While there are a plethora of methods for multi-agent trajectory forecasting \cite{RudenkoPalmieriEtAl2019}, many are developed without explicit consideration for downstream use cases and only produce future agent tracklets as prediction outputs. As a result, state-of-the-art trajectory forecasting methods are rarely used by the wider robotics community, especially due
to the complexities of incorporating multimodal prediction distributions (commonly produced by current generative trajectory forecasting methods). 
Recently, works such as the Neural Motion Planner (NMP) \cite{ZengLuoEtAl2019} have designed output representations for trajectory forecasting that allow for more direct integration with downstream motion planners. In particular, NMP's perception module outputs a cost volume over time that is used with a sampling-based motion planner to evaluate trajectory samples at runtime. 
However, while cost volumes are amenable for downstream planning methods, such as sampling-based motion planners, the predictions do not account for the environment's dynamic response to the ego-vehicle's actions. In this work, we present a trajectory forecasting representation that allows for the direct optimization of the ego-vehicle's future controls while simultaneously predicting the trajectories of other agents. In particular, the predictions take into consideration the ego-vehicle's future motion as well as agent-agent interactions.

To incorporate trajectory forecasting in planning, motion planners usually either optimize a plan in position space that avoids predicted agent locations (using a lower-level feedback controller to track the plan) \cite{ZieglerBenderEtAl2014,LiuLeeEtAl2017} or project the predictions into the ego-vehicle's control space and then optimize its controls (which are directly executed) \cite{FanZhuEtAl2018}. 
Our proposed planner takes a similar approach to the latter, 
and leverages the future-conditional \algname{} predictions produced by our model to produce interaction-aware plans.

\textbf{Dynamical System Representations of Multiple Agents.} One of the earliest trajectory forecasting methods,
the Social Force model \cite{HelbingMolnar1995}, reasons about inter-agent interactions in terms of explicit physical forces, taking a dynamical system view of human interactions. More recently, works have studied how general dynamical system models of environments can be learned from observations. For instance, \cite{WatterSpringenbergEtAl2015,HafnerLillicrapEtAl2019,IchterPavone2019} learn a latent dynamics model directly from sensor input that is then combined with standard sampling-based motion planners or trajectory optimization algorithms to move an ego-agent through an environment. Such ideas have even been applied to trajectory forecasting; Neural Relational Inference \cite{KipfFetayaEtAl2018} first predicts the latent dynamics of agent interactions (i.e., which agents are interacting) and then forecasts their trajectories taking the predicted interactions into account. Our work differs in that it produces an \textit{explicit} physical (not latent) dynamical system model which can then be used to predict the future states of other agents. This is beneficial as it leads to an easier integration with downstream planning and control algorithms as well as being more immediately interpretable since all values are in the original (not latent) state space.

\section{Problem Formulation}\label{sec:problem_formulation}
We aim to generate future trajectory distributions for a time-varying number $N(t)$ of interacting agents $a_1,...,a_{N(t)}$. Each agent $a_i$ has a semantic class $S_i$, e.g., Car, Bus, Pedestrian, and state $\mathbf{s}_i^{(t)} \in \mathbb{R}^{D_i}$ at time $t$. We denote $\mathbf{s}^{(t)} \in \mathbb{R}^{F}$ as the concatenation of all agent states where $F = \sum_{i=1}^{N(t)} D_i$ is the full state dimension of the system at time $t$. Given the states $\mathbf{s}_i^{(t)}$ for each agent $a_i$ and their history for the previous $H$ timesteps, denoted as $\mathbf{x} = \mathbf{s}^{(t - H : t)} \in \mathbb{R}^{(H + 1) \times F}$, we seek a distribution over all agents' future states for the next $T$ timesteps, denoted as $p(\mathbf{y} \mid \mathbf{x})$ where $\mathbf{y} = \mathbf{s}^{(t + 1 : t + T)} \in \mathbb{R}^{T \times F}$. Since one of our key desiderata is to enable the close integration of trajectory prediction with downstream planning and control, we also condition on the ego-agent's future motion plan, $\mathbf{u}_\text{R} \in \mathbb{R}^{T \times C}$, where $\mathbf{u}_\text{R}^{(t)} \in \mathbb{R}^C$ is the ego-agent's control at time $t$, to obtain $p(\mathbf{y} \mid \mathbf{x}, \mathbf{u}_\text{R})$.
Such information is readily available online, e.g., when evaluating responses to a set of motion primitives or finding the best future motion plan via optimization (as in \cref{sec:planning}). Throughout this work, the terms ``ego-agent," ``ego-robot," and ``ego-vehicle" refer to the robot being controlled.

\section{Trajectory Forecasting with \algname{}}\label{sec:forecasting}
Affine Time-Varying (ATV) systems generalize linear time-varying systems and are well-studied in planning, decision making, and control \cite{Luenberger1979}. They are commonly used to model the joint evolution of agent states (e.g., position, velocity) over time. 
We choose this specific representation as (i) linear systems are straightforward to integrate with planning and control, yielding computationally-efficient algorithms, (ii) time-variance allows the (otherwise linear) model to capture non-linear effects, and (iii) the affine term accounts for the linearization of non-linear agent dynamics.

Illustrated in \cref{fig:architecture}, our method produces ATV systems by first representing agents and their interactions as nodes and edges in a directed complete spatiotemporal graph, $G = (V, E)$.
Each edge is then modeled with a deep recurrent probabilistic encoder-decoder architecture that takes as input the state history of the connected nodes as well as the ego-robot's future motion plan and produces a set of ATV system submatrices. The matrices are then stacked together as in \cref{fig:architecture} to form an overall ATV system that describes the future motion of all agents. We will now dive into the details of each component.

\textbf{Encoding Edges.} A directed edge from agent $a_i$ to agent $a_j$, $(a_i, a_j) \in E$, is modeled using two Long Short-Term Memory (LSTM) networks \cite{HochreiterSchmidhuber1997}, each with $32$ hidden dimensions. Agent $a_j$'s state history $\mathbf{s}^{(t-H:t)}_j$ is directly encoded by an LSTM whereas $a_i$'s state history is first made relative to $a_j$'s current state, $\mathbf{s}^{(t-H:t)}_i - \mathbf{s}^{(t)}_j$, before being encoded by another LSTM. Encoding edges this way ensures that they are modeled asymmetrically. The LSTM's last hidden states are concatenated to form a backbone representation vector, $\mathbf{e}$.

\textbf{Encoding the Ego-Agent's Future.} Producing predictions that are cognizant of future ego-agent motion is an important capability for trajectory prediction systems. For example, it allows one to evaluate candidate motion plans while taking into account potential responses from other agents. Our model encodes the future $T$ timesteps of the ego-agent's motion plan, $\mathbf{u}_\text{R}$, using a bi-directional LSTM with 32 hidden dimensions. A bi-directional LSTM is used due to its strong performance on other sequence summarization tasks \cite{BritzGoldieEtAl2017}. The final hidden states are then concatenated into $\mathbf{e}$.

\textbf{Modeling Multimodality.}
A key challenge in modeling human behavior is their inherent multimodality (i.e., the potential for many possible future actions). To capture this, our model produces a \textit{mixture} of ATV systems, where each component ATV system models a high-level latent behavior for the scene (e.g., groups of vehicles slowing down for a red traffic light). We use a CVAE~\cite{SohnLeeEtAl2015} with a discrete Categorical latent variable $z \in Z$ to decompose the desired $p(\mathbf{y} \mid \mathbf{x}, \mathbf{u}_\text{R})$ distribution as $p(\mathbf{y} \mid \mathbf{x}, \mathbf{u}_\text{R}) = \sum_z p(\mathbf{y} \mid z, \mathbf{x}, \mathbf{u}_\text{R})\, p(z \mid \mathbf{x}, \mathbf{u}_\text{R})$. In essence, $|Z|$ individually-predicted ATV system distributions $p(\mathbf{y} \mid z, \mathbf{x}, \mathbf{u}_\text{R})$ are combined in a weighted sum, according to their mode likelihoods $p(z \mid \mathbf{x}, \mathbf{u}_\text{R})$, to form a single \algname{} model $p(\mathbf{y} \mid \mathbf{x}, \mathbf{u}_\text{R})$. Note that the latent variable $z$ defines a mode for the \textit{entire} predicted ATV system, which aids in producing self-consistent predictions, in contrast to prior CVAE-based approaches that produce $z$ independently per agent~\cite{LeeChoiEtAl2017,IvanovicPavone2019,SalzmannIvanovicEtAl2020}.

\textbf{Generating \algname{}: Combining Dynamics and Learning.} The latent variable $z$ and the backbone representation vector $\mathbf{e}$ are then fed into the decoder, a 128-dimensional Gated Recurrent Unit (GRU)~\cite{ChoMerrienboerEtAl2014}. Each GRU cell outputs a set of submatrices, $A_{z, ij}^{(t)},\ B_{z, i}^{(t)},\ Q_{z, i}^{(t)}$, for each prediction timestep $t$ and mode $z$. The matrices are then stacked together as in \cref{fig:architecture} to form the following overall ATV system that describes the future motion of all agents for a specific mode, %
\begin{align}
    \mathbf{s}_z^{(t + 1)} &= A_z^{(t)} \mathbf{s}_z^{(t)} + B_z^{(t)} \mathbf{u}^{(t)}_{\text{R}} + \mathbf{c}_z^{(t)} + Q_z^{(t)} \mathbf{w}^{(t)}, \label{eqn:uncertain_ATV}%
\end{align}
where $\mathbf{w}^{(t)} \sim \mathcal{N}(\mathbf{0}, I)$. \cref{eqn:uncertain_ATV} is linear Gaussian as its only source of uncertainty is $w^{(t)}$. Thus, each component distribution $p(\mathbf{y} \mid z, \mathbf{x}, \mathbf{u}_\text{R})$ can be obtained by iteratively evaluating
\cref{eqn:uncertain_ATV}.

As illustrated in \cref{fig:architecture}, our model leverages individual agent dynamics as an inductive bias and only produces submatrices for system components that are not already modeled by dynamics, e.g., the off-diagonal blocks of $A$ which account for inter-agent interactions. Specifically, agent $a_j$'s effect on agent $a_i$ is contained in $A_{ij} \in \mathbb{R}^{D_i \times D_j}$.
Our model leverages the semantic class $S_i$ of each agent $a_i$ to fill its corresponding block in $A$ and $\mathbf{c}$ with the agent's corresponding dynamical system matrices. For example, we model pedestrians as double integrators in this work, thus if agent $a_i$ was a pedestrian then the $A_{ii} \in \mathbb{R}^{D_i \times D_i}$ submatrix would be the double integrator dynamics matrix \cite{Luenberger1979} and $\mathbf{c}_i$ would be zero. Note that the ego-agent's row in $A$ only contains its own dynamics ($A_{11}$ in \cref{fig:architecture}), while the rest of the row consists of zero matrices. This reflects the fact that the ego-agent is only influenced by its controls, as determined by a downstream planner. Accordingly, the ego-agent's entry in $B$ is governed by dynamics while the rest of the blocks in $B$ are produced by the model, representing the effect of the ego-agent's control on others' behavior.

As can be seen in \cref{eqn:uncertain_ATV}, an ATV system is comprised of an autonomous term ($A \in \mathbb{R}^{F \times F}$ specifies the evolution of the system in the absence of ego-agent controls), a controlled term ($B \in \mathbb{R}^{F \times C}$ determines how the ego-agent's controls affect other agents), an affine term ($\mathbf{c} \in \mathbb{R}^{F}$ arises from linearizing non-linear dynamics), and an uncertainty term ($Q \in \mathbb{R}^{F}$ specifies the model's uncertainty). 
At a high level, we consider the ego-agent's future motion plan $\mathbf{u}_\text{R}$ as the ``control" of the ATV system, which can then be optimized over using classical optimal control techniques. 

Note that our model naturally handles a time-varying number of agents. Past trajectories with various lengths can be naturally encoded in LSTMs
and predictions are made assuming the agents present at time $t$ will be present until time $t+T$. The graph is recreated every timestep from the present agents.

\textbf{Training Objective.}
We train the model by maximizing the following discrete InfoVAE objective function \cite{ZhaoSongEtAl2019,SalzmannIvanovicEtAl2020}, 
\begin{equation}\label{eqn:loss_fn}
\begin{aligned}
\mathbb{E}&_{z \sim q(\cdot \mid \mathbf{x}, \mathbf{y}, \mathbf{u}_\text{R})} \big[\log p(\mathbf{y} \mid z, \mathbf{x}, \mathbf{u}_\text{R})\big] - \beta D_{KL}\big(q(z \mid \mathbf{x}, \mathbf{y}, \mathbf{u}_\text{R}) \parallel p(z \mid \mathbf{x}, \mathbf{u}_\text{R})\big) + \alpha I_{q} (\mathbf{x}; z).
\end{aligned}
\end{equation}
where $q(z \mid \mathbf{x}, \mathbf{y}, \mathbf{u}_\text{R})$ is only produced during training from a bi-directional LSTM with 32 hidden dimensions that encodes each agent’s future trajectory. To keep the number of parameters low, we use extensive weight-sharing throughout the model. For example, all LSTMs that encode state histories for the same agent type (e.g., vehicles) will share weights.

\section{Incorporating \algname{} Predictions in Motion Planning}\label{sec:planning}

The ATV form of the overall system dynamics was chosen explicitly because it can be easily integrated within existing planning algorithms while being expressive enough to accurately model agent-agent interactions. In this section, we propose one such method that extends standard Model Predictive Control (MPC)~\cite{RawlingsMayneEtAl2017} to utilize the multimodal \algname{} predictions described in the previous section and produce an interaction-aware motion plan for the ego-robot. We first provide a brief background on MPC and then describe our proposed planning method.

\subsection{Model Predictive Control}
MPC is an optimal control framework which casts an optimal control problem as an optimization over a finite-horizon control sequence to minimize a designer-specified objective function while satisfying dynamics and control constraints. The optimization problem over a horizon of $T$ steps for discrete-time systems in our context can be written in the form
\begin{align}\label{eqn:generic_mpc}
\begin{split}
\min_{\mathbf{\bar{q}},\mathbf{\bar{u}}_{\text{R}}} \quad h(\mathbf{q}^{(T)}) + \sum_{t=1}^{T-1}g(\mathbf{q}^{(t)},\mathbf{u}_{\text{R}}^{(t)}) \quad
\textrm{s.t.} \quad &\mathbf{q}^{(t+1)} = f(\mathbf{q}^{(t)}, \mathbf{u}_{\text{R}}^{(t)}, t)\\ 
                    &\mathbf{q}^{(1)} = \mathbf{q_{\textrm{init}}},\ \mathbf{q}^{(t)} \in \mathcal{Q}^{(t)}, \ \mathbf{u}_{\text{R}}^{(t)} \in \mathcal{U}^{(t)}.
\end{split}
\end{align}
In this section, we use $\mathbf{q}^{(t)} \in \mathbb{R}^n$ and $\mathbf{u}_{\text{R}}^{(t)} \in \mathbb{R}^m$ to represent the ego-robot system state and control input at timestep $t$, and define $\mathbf{\bar{q}} = (\mathbf{q}^{(1)}, \ldots, \mathbf{q}^{(T)})$ and $\mathbf{\bar{u}}_{\text{R}} = (\mathbf{u}_{\text{R}}^{(1)}, \ldots, \mathbf{u}_{\text{R}}^{(T-1)})$. \cref{eqn:generic_mpc} assumes that the robot is initially at state $\mathbf{q_\textrm{init}}$ and obeys dynamics given by $f: \mathbb{R}^n \times \mathbb{R}^m \mapsto \mathbb{R}^n$. In the objective function we have $h: \mathbb{R}^n \mapsto \mathbb{R}$ which yields the terminal state cost and $g: \mathbb{R}^n \times \mathbb{R}^m \mapsto \mathbb{R}$ which yields the stage cost. The admissible state set and control set at each time instant are respectively denoted by $\mathcal{Q}^{(t)}\subset \mathbb{R}^n$ and $\mathcal{U}^{(t)}\subset \mathbb{R}^m$.

The MPC framework entails solving \cref{eqn:generic_mpc} in a receding-horizon fashion. At each iteration, the open-loop control sequence is computed and the first action is applied. At the next timestep, the new optimization is solved from the new system state and this process repeats until some termination condition is achieved.

\subsection{Planning with \algname{}}
We propose a motion planning strategy based on MPC that leverages the \algname{} trajectory forecasting representation to plan smooth, socially-aware trajectories inspired by Contingency MPC~\cite{AlsterdaBrownEtAl2019}. Specifically, we extend the formulation given in \cref{eqn:generic_mpc} from planning a single open-loop control sequence to optimizing multiple control sequences simultaneously, with each sequence accounting for a particular evolution mode of the agents in a scene (determined by the latent variable $z$). These ``parallel" control sequences are constrained to achieve agreement across the first $t_c$ timesteps to make the resulting output actionable, which we refer to as the consensus horizon. We formulate the corresponding optimization as 
\begin{align}\label{eqn:consensus_mpc}
\begin{split}
\min_{\mathbf{\bar{s}},\mathbf{\bar{u}}_{\text{R}}} \quad &\sum_{z \in Z}h(\mathbf{s}_z^{(T)}) + \sum_{z \in Z}\sum_{t=1}^{T-1}g(\mathbf{s}_z^{(t)}, \mathbf{u}_{\text{R}, z}^{(t)}) \\
\textrm{s.t.} \quad &\mathbf{s}_z^{(t + 1)} = A_z^{(t)} \mathbf{s}_z^{(t)} + B_z^{(t)} \mathbf{u}^{(t)}_{\text{R}, z} + \mathbf{c}_z^{(t)}\ \forall\ z \in Z\\ 
&\mathbf{s}_z^{(1)} = \mathbf{s_{\textrm{init}}},\ \mathbf{s}_z^{(t)} \in \mathcal{S}^{(t)},\ \mathbf{u}_{\text{R}, z}^{(t)} \in \mathcal{U}^{(t)}\ \forall\ z \in Z\\
&\mathbf{u}_{\text{R}, z_1}^{(t)} = \mathbf{u}_{\text{R}, z_2}^{(t)}\ \forall\ z_1, z_2 \in Z,\ t \in \{1,2,\ldots,t_c\}.
\end{split}
\end{align}
This formulation remains largely similar to the one in \cref{eqn:generic_mpc}, but we now extend the system of interest from just the ego-robot to the set of all agents in a scene. The system state, as in \cref{sec:problem_formulation}, is represented by $\mathbf{s}$ which is the concatenation of all agent states in a scene, including that of the ego-robot, $\mathbf{q}$. Notably, we use \cref{eqn:uncertain_ATV} as the system dynamics constraints, which naturally includes the robot vehicle's deterministic dynamics. However, control of the system may only be achieved through those of the robot vehicle, $\mathbf{u}_{\text{R}}$, as in \cref{eqn:generic_mpc}. Note that states, controls and sets are now indexed by the latent variable $z \in Z$ which references a particular mode's prediction. As such, the decision variables for this problem are $\mathbf{\bar{s}} = \bigcup_{z \in Z} (\mathbf{s}_z^{(1)}, \ldots, \mathbf{s}_z^{(T)})$ and, slightly abusing the notation, $\mathbf{\bar{u}_R} = \bigcup_{z \in Z} (\mathbf{u}_{\text{R},z}^{(1)}, \ldots, \mathbf{u}_{\text{R},z}^{(T-1)})$, representing the state and control trajectories across all modes indexed by $z \in Z$. 

By planning in this manner, each parallel horizon only needs to account for a specific prediction mode rather than multiple prediction modes. Only the states and controls within the consensus horizon are subject to constraints over $z$; however, in the near-term it is unlikely that significant differences in the environment state $\mathbf{s}_z$ would emerge. The consensus horizon serves to ensure that the ego-robot will always take an action that can accommodate any of the predicted future modes. In this way, we posit that this approach should reduce conservatism in planning without needing to sacrifice constraints. 

\textbf{Benefits of \algname{} for Planning.} The key advantage of planning with \algname{}, rather than agent tracklets, is that the optimizer can use the dynamical system representation to fully explore the space of robot controls around the nominal robot future. Notably, it can observe how these controls affect the distribution of other agents responses yielding a ``dynamic" prediction, i.e., one that changes post-hoc with different control $\mathbf{u}_\text{R}$. Although agent tracklets can also be conditioned on a nominal robot future, the resulting predictions are ``static" and cannot change post-hoc. With agent tracklets, one would need to query the prediction model for every tentative $\mathbf{u}_\text{R}$, which may not be feasible for real-time robotic use cases.

\textbf{More Sophisticated Probabilistic Reasoning.} In its current form, \cref{eqn:consensus_mpc} weighs the predictions of the $z$ modes equally. However, there exist more sophisticated ways of utilizing the set of \algname{} predictions for planning. In particular, the prediction model provides information about mode likelihood $p(z \mid \mathbf{x}, \mathbf{u}_\text{R})$ and state uncertainty $Q_z$. One can reason about the relative likelihood of each mode and accordingly weigh the planning costs in \cref{eqn:consensus_mpc}. Such a weighting would allow the planner to prioritize planning for situations that are more likely to occur, while placing lower importance on low-likelihood predictions. Furthermore, state uncertainty information could inform collision avoidance bounds for state constraints (e.g., as chance constraints). Further investigation as to how this probabilistic information can be used is left as future work.

\textbf{Safety Considerations.} The proposed planner is intended to promote efficient, prescient navigation for autonomous vehicles. As a result, it does not currently guarantee safe behavior, given that we are reasoning about uncertain predictions of other agents' future behaviors -- a fact that is true for virtually every probabilistic planner. Further, feasibility is also not guaranteed since we consider a finite horizon. However, the enforcement of hard safety constraints could be achieved by pairing our MPC planner with a lower-level, reachability-based controller (e.g.,~\cite{LeungSchmerlingEtAl2019}). In fact, the structure of our MPC algorithm simplifies the integration of these two levels of the control hierarchy, as \cref{eqn:consensus_mpc} permits the inclusion of lower-level reachability terms or constraints.

\section{Experimental Results} 

We evaluate our method in two scenarios, an illustrative charged particle system and the nuScenes dataset~\cite{CaesarBankitiEtAl2019}. The charged particle system serves as a controlled experiment with well-known non-linear system dynamics, and the nuScenes dataset evaluates our approach's feasibility for modeling and planning with real-world pedestrians and vehicles. We demonstrate the viability of \algname{} as a trajectory forecasting representation and find that it significantly reduces planning complexity.
Our trajectory forecasting model and motion planner were implemented in PyTorch and Julia, respectively. We trained the model on a Ubuntu 18.04 desktop computer containing an AMD Ryzen 1800X CPU and two NVIDIA GTX 1080 Ti GPUs for 100 epochs on the particle dataset ($|Z| = 25$) and 16 epochs on the nuScenes dataset ($|Z| = 5$). More training details can be found in the appendix.

\subsection{Two Particle System}

\begin{figure}[t]
    \centering
    \includegraphics[width=0.25\linewidth]{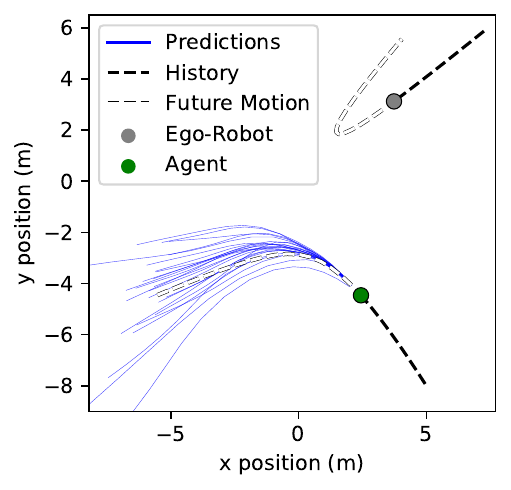}
    \includegraphics[width=0.74\linewidth]{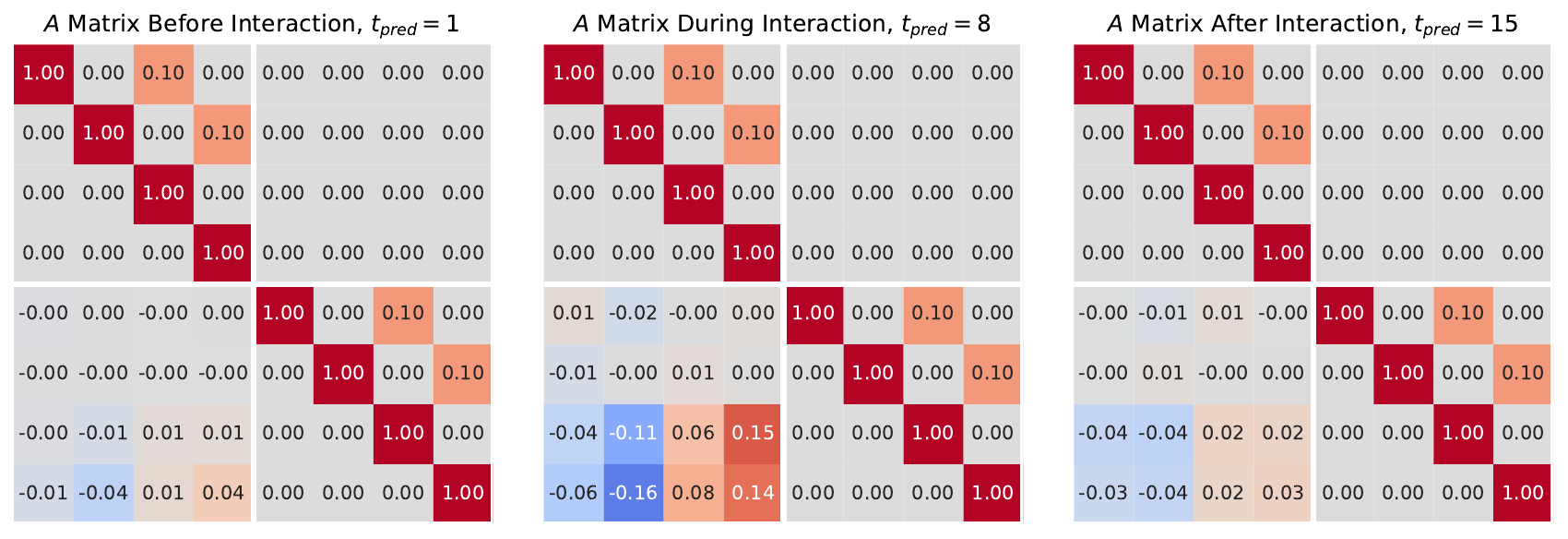}
    \caption{\textbf{Left:} Our mixture model's predictions for an agent interacting with the ego-robot. Each blue line corresponds to one Affine Time-Varying system component. \textbf{Right:} Our model's most likely output $A_z$ matrix across prediction timesteps. The joint state $\mathbf{s}_{1,2}$ is the concatenation of the ego-robot and agent states, hence why the top-right submatrix is all zeros.
    }
    \label{fig:particle}
    \label{fig:particle_A_matrices}
\end{figure}

To show that the \algname{} representation can model non-linear dynamics, we consider a scenario where two agents following double-integrator dynamics interact according to the Social Forces model~\cite{HelbingMolnar1995}, visualized in \cref{fig:particle}. One of the agents, denoted as the ``ego-robot", acts according to a given set of control actions $\mathbf{u}_\text{R}$. The other, designated as the ``agent", has an initial velocity $v_0 \in [4, 12]$m/s and avoids the ego-robot with a force proportional to the inverse-square of the distance between the two (a non-linear relation). We generated a dataset of 1000 $3$s-long scenes for training and tested on 200 held-out scenes. The model was trained to predict the next 12 timesteps (1.2s) having observed the previous 8 timesteps (0.8s).

\textbf{Interpreting $A$.} As can be seen in \cref{fig:particle_A_matrices}, the lower-left block of the most likely $A_z$ matrix (corresponding to the ego-robot's effect on the agent) has nonzero components that appear during the interaction and dissipate after. This has a direct interpretation in this scenario, as the bottom two rows affect the agent's velocity along the $x$ and $y$ directions $(v_x, v_y)$ for the next timestep. In particular, as the ego-robot moves faster in the negative $x$ and $y$ directions (towards the agent), the agent's next $v_x, v_y$ values would decrease, i.e., the agent would move away.

\subsection{nuScenes Dataset}
\begin{figure}[t]
    \centering
    \includegraphics[width=0.29\linewidth]{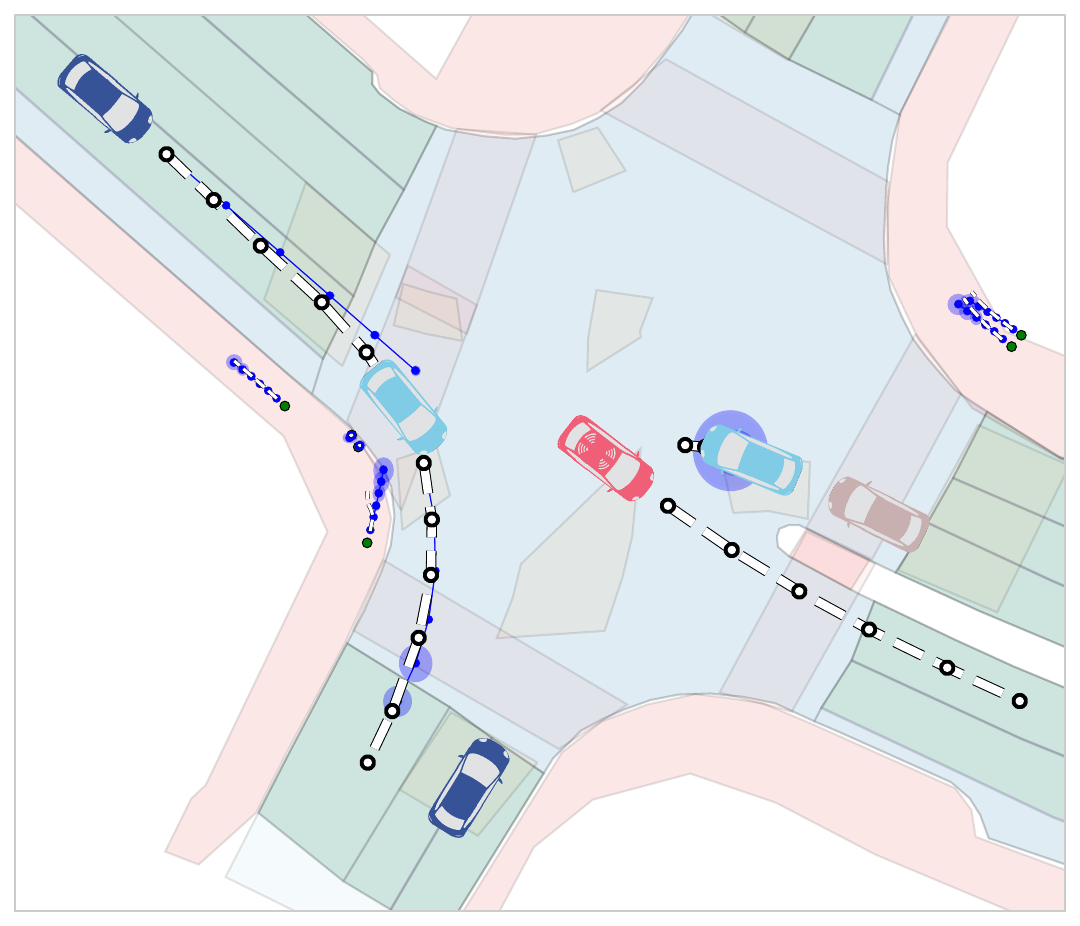}
    \includegraphics[width=0.29\linewidth]{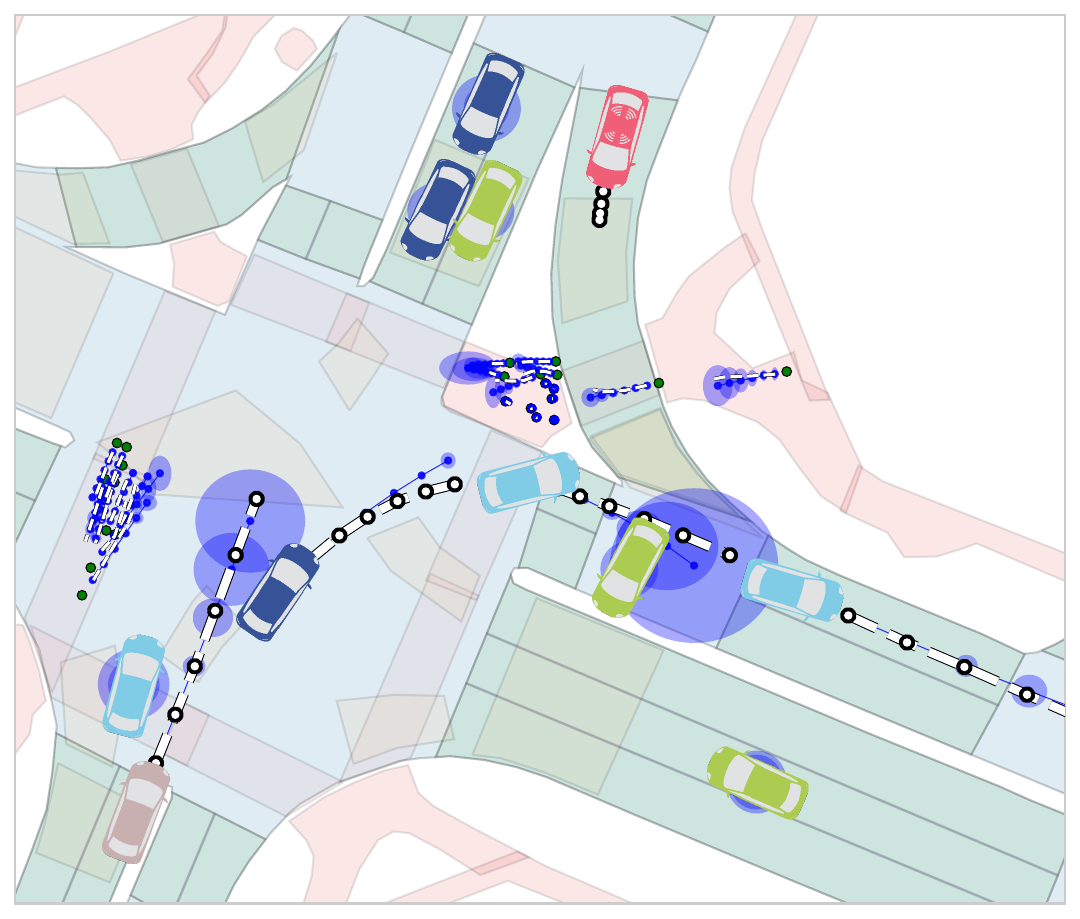}
    \includegraphics[width=0.29\linewidth]{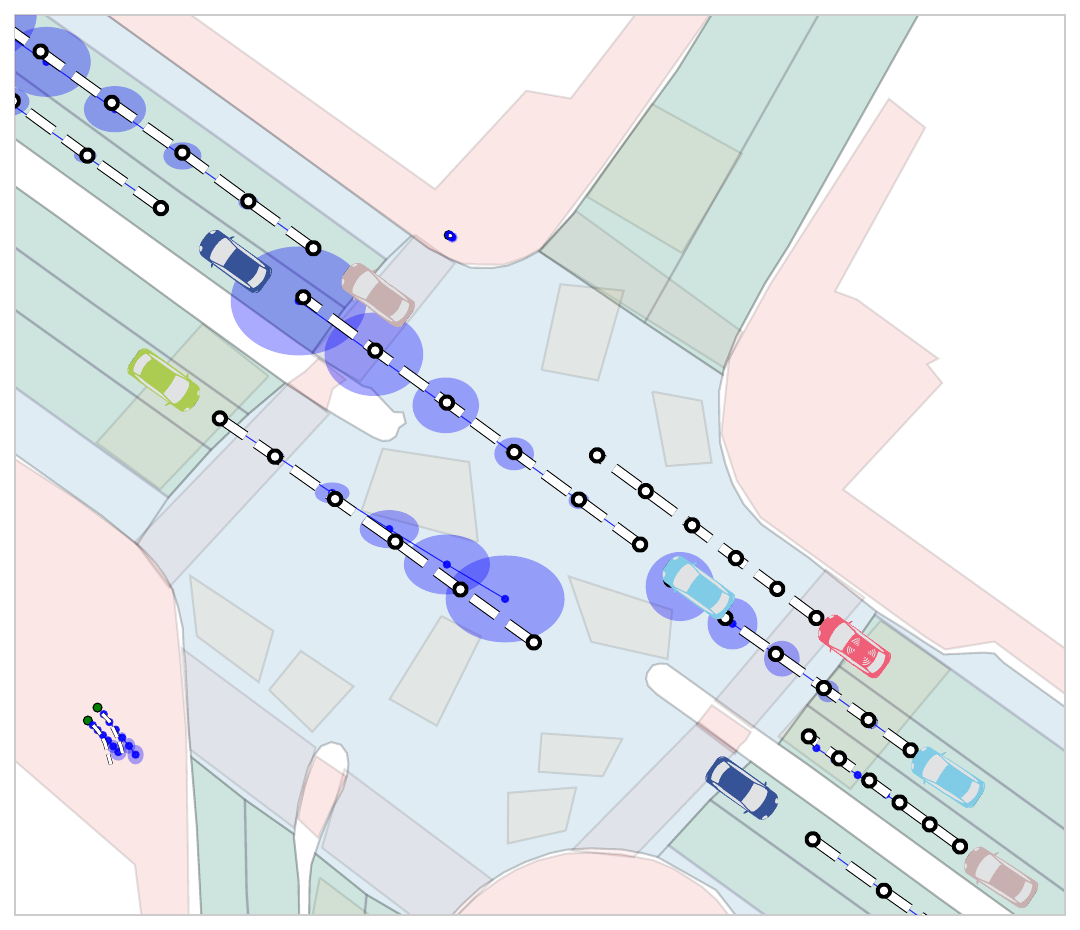}
    \raisebox{0.3\height}{\includegraphics[width=0.09\linewidth]{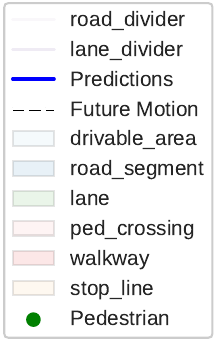}}
    \caption{Our method's predictions (blue) on various scenes from the nuScenes dataset. The ego-vehicle is shown in red. Ellipsoids show 95\% probability. For clarity, only predictions from the most likely $z$ mode are shown.}
    \label{fig:nuscenes}
\end{figure}

nuScenes \cite{CaesarBankitiEtAl2019} is a large-scale real-world dataset for autonomous driving comprised of 1000 20s-long scenes in Boston and Singapore. 
We model vehicles as dynamically-extended unicycles \cite{LaValle2006BetterUnicycle} and pedestrians as double integrators\footnote{While recent work \cite{SalzmannIvanovicEtAl2020} models humans as single integrators, we choose a double integrator so that pedestrians can have autonomous motion even if $\mathbf{u}_\text{R} = \mathbf{0}$ (a single integrator only moves with nonzero controls).}.
Notably, this means that we linearize vehicles' dynamics about their current state to obtain linear dynamical system matrices for their corresponding ATV system components. 
\cref{fig:nuscenes} shows a set of predictions from our model. In it, we can see that the model makes smooth predictions with uncertainties that sensibly grow in time.

\begin{table}[t]
\centering
\caption{FDE for (a) vehicles and (b) pedestrians across time for our method's most likely mode compared to state-of-the-art models on the nuScenes dataset. 
Starred methods had $22$-$24$cm subtracted from their reported values (their detection/tracking error~\cite{CasasGulinoEtAl2019}), as we do not use a detector/tracker. Our model lags slightly in raw prediction error (due to its affine output structure), but is more amenable to and yields significant computational efficiency improvements for downstream planning and control.
}
\begin{tabular}{l|ccc}
\toprule
\multicolumn{4}{c}{\textbf{(a) Vehicle FDE (m)}}\\
\midrule
\textbf{Method} & @1s & @2s & @3s \\
\midrule
S-LSTM$^*$~\cite{AlahiGoelEtAl2016,CasasGulinoEtAl2019} & $0.47$ & - & $1.61$\\
CSP$^*$~\cite{DeoTrivedi2018,CasasGulinoEtAl2019} & $0.46$ & - & $1.50$\\
CAR-Net$^*$~\cite{SadeghianLegrosEtAl2018,CasasGulinoEtAl2019} & $0.38$ & - & $1.35$\\
SpAGNN$^*$~\cite{CasasGulinoEtAl2019} & $0.36$ & - & $1.23$\\
Ours & $0.26$ & $0.98$ & $2.20$\\
Trajectron++~\cite{SalzmannIvanovicEtAl2020} & $0.07$ & $0.45$ & $1.14$\\
\bottomrule
\end{tabular}
\hfill%
\begin{tabular}{l|ccc}
\toprule
\multicolumn{4}{c}{\textbf{(b) Pedestrian FDE (m)}}\\
\midrule
\textbf{Method} & @1s & @2s & @3s \\
\midrule
Ours & $0.06$ & $0.21$ & $0.42$\\
Trajectron++~\cite{SalzmannIvanovicEtAl2020} & $0.01$ & $0.17$ & $0.37$\\
\bottomrule
\end{tabular}
\label{tab:nuscenes_fde}
\end{table}

\textbf{Quantitative Performance.} We compare our trajectory forecasting method against an array of state-of-the-art approaches using the Final Displacement Error (FDE) metric, i.e., the $\ell_2$ distance between the prediction and ground truth trajectories at the last prediction timestep. Our model is compared against LSTM-based \cite{AlahiGoelEtAl2016,DeoTrivedi2018,SadeghianLegrosEtAl2018} and graph neural network-based \cite{SalzmannIvanovicEtAl2020,CasasGulinoEtAl2019} approaches that use a variety of pooling operations to encode agent-agent interactions and scene context (e.g., semantic maps). 

The results are summarized in \cref{tab:nuscenes_fde}, and show that our method is similar in performance to Trajectron++~\cite{SalzmannIvanovicEtAl2020} for pedestrians, and better than most for early timesteps on vehicles, with performance dropping off as expected in later timesteps. As mentioned earlier, the core reason for this is that our method is significantly more structured than other models (namely, it is affine). In comparison, the baselines all produce predictions directly from their decoder networks and are optimized to minimize positional prediction error. In contrast, our method predicts each agent's \textit{full state}, e.g., position, velocity, and heading for vehicles. As a result, pure positional prediction performance is not a key factor in evaluating our method, especially due to the improved efficiency of downstream planning. 

\begin{figure}[t]
    \centering
    \includegraphics[width=0.295\linewidth]{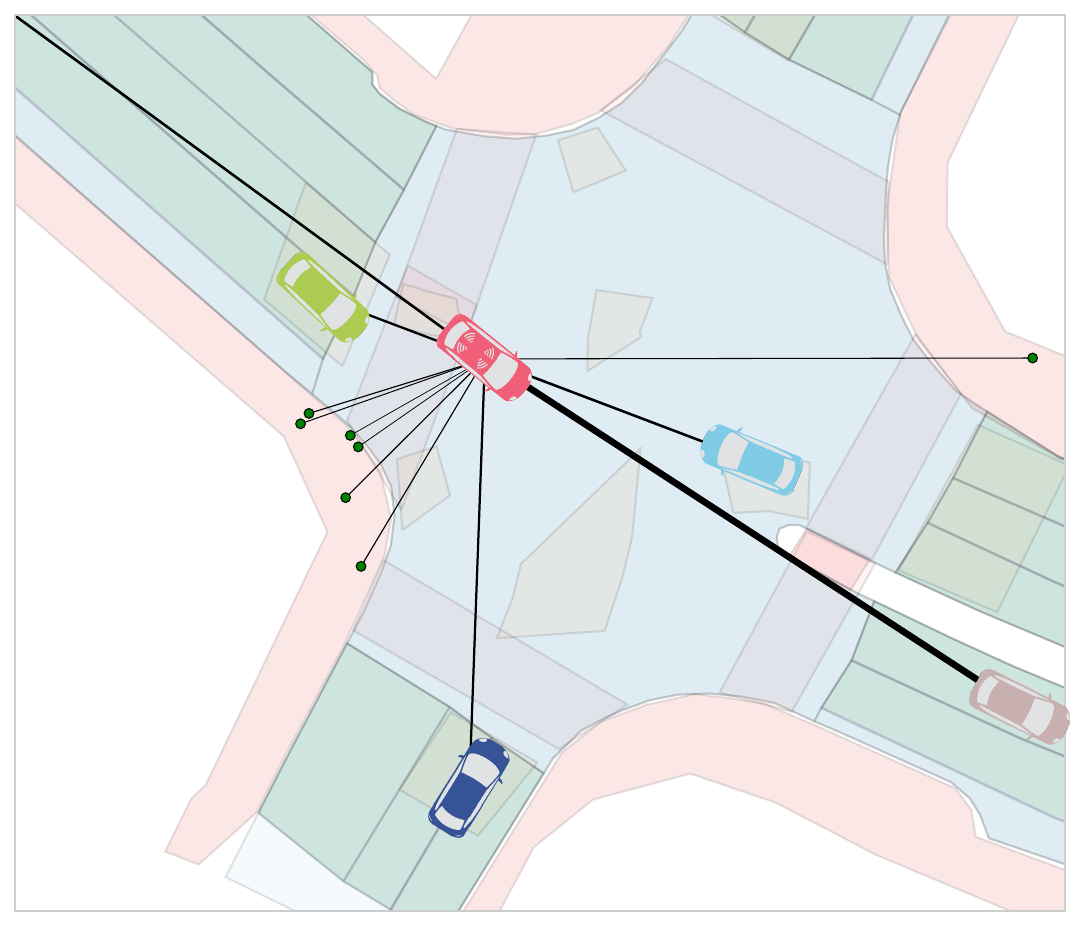}
    \includegraphics[width=0.295\linewidth]{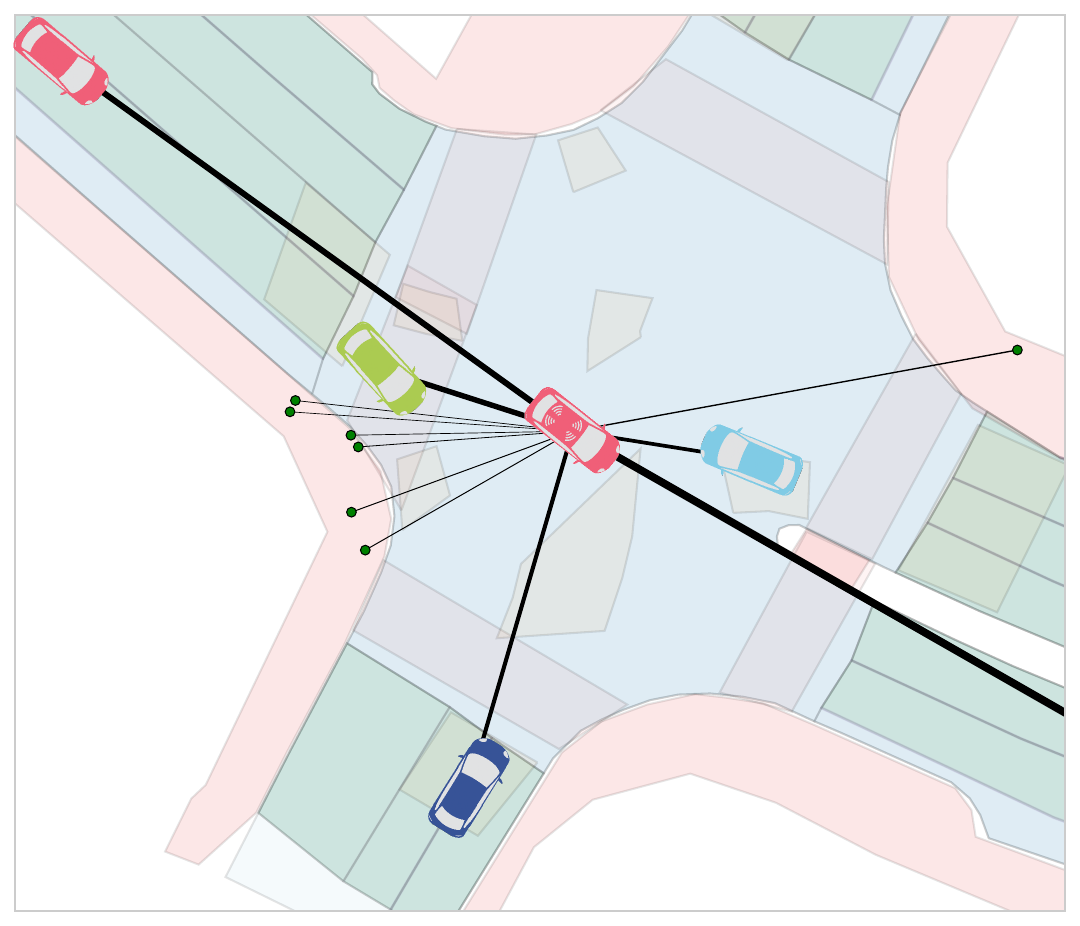}
    \includegraphics[width=0.295\linewidth]{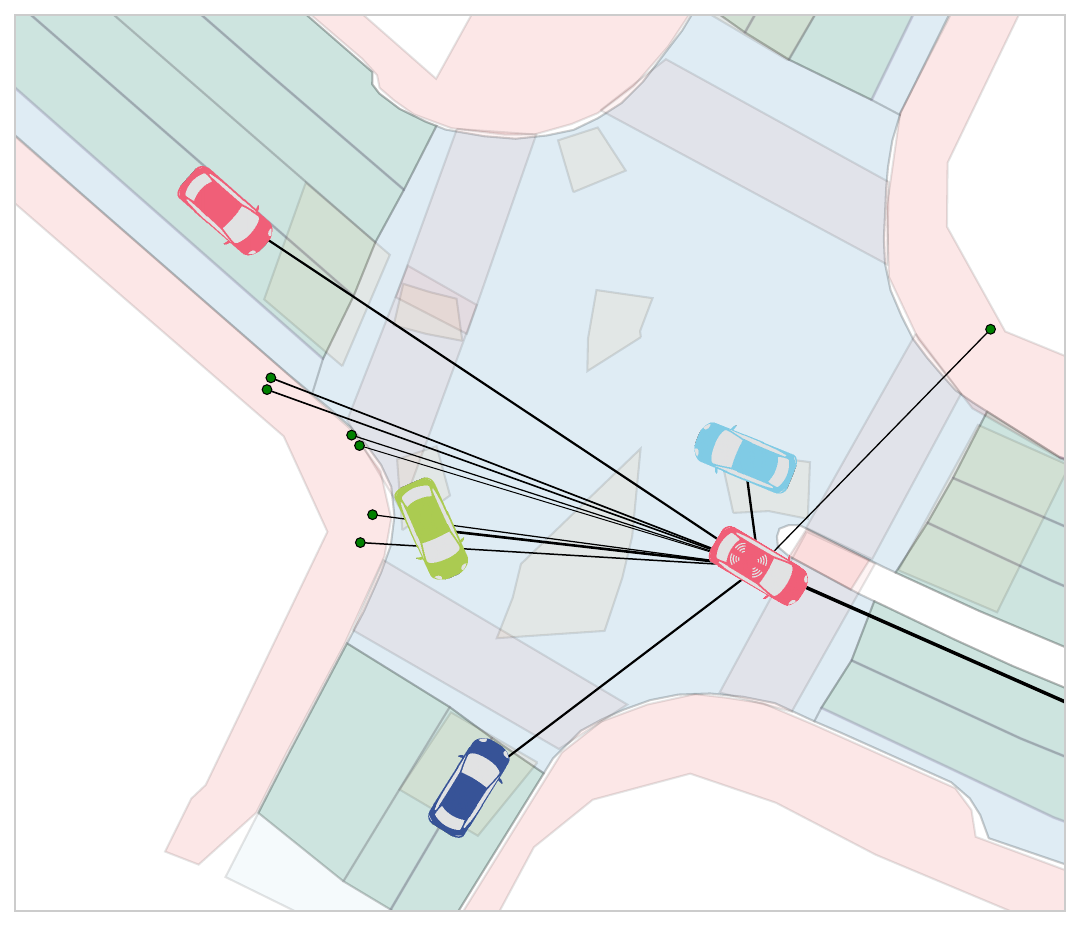}
    \raisebox{0.4\height}{\includegraphics[width=0.095\linewidth]{figures/legend.pdf}}
    \caption{The $B$ matrix allows us to infer the influence that the ego-agent (red vehicle crossing the intersection) has on other agents. Line thickness is determined by the Frobenius norm of the corresponding block in the $B_z$ matrix of the most-likely mode $z$. As can be seen, when the ego-vehicle is in the middle of the intersection it is maximally influencing the other vehicular agents, and less so before and after being in the intersection. Pedestrians are equally influenced through time.}
    \label{fig:nuscenes_B_graph}
\end{figure}

\textbf{Inferring Ego-Vehicle Influence.} 
The block components of the $B$ matrix dictate the effect of the ego-robot's control on the other agent's predicted states. Thus, by analyzing the $B$ matrix one can determine how other agents will be affected by the ego-robot. \cref{fig:nuscenes_B_graph} visualizes one such analysis, where the strength of the connection from the ego-robot to a specific agent $a_i$ at time $t$ is proportional to the Frobenius norm of $B_{z,i}^{(t)}$. Clear temporal patterns can be seen in the connections between the ego-vehicle and the vehicles in the scene. In particular, their $B_{z,i}^{(t)}$ norms are 
the highest when the ego-vehicle is in the middle of the intersection, indicating that this is a time of maximal influence on the other vehicles. Additional figures of the corresponding $B_z$ matrices can be found in the appendix.

\subsection{Interaction-Aware Motion Planning}\label{sec:planning_exp}

\begin{figure}[t]
    \centering
    \includegraphics[width=0.29\linewidth]{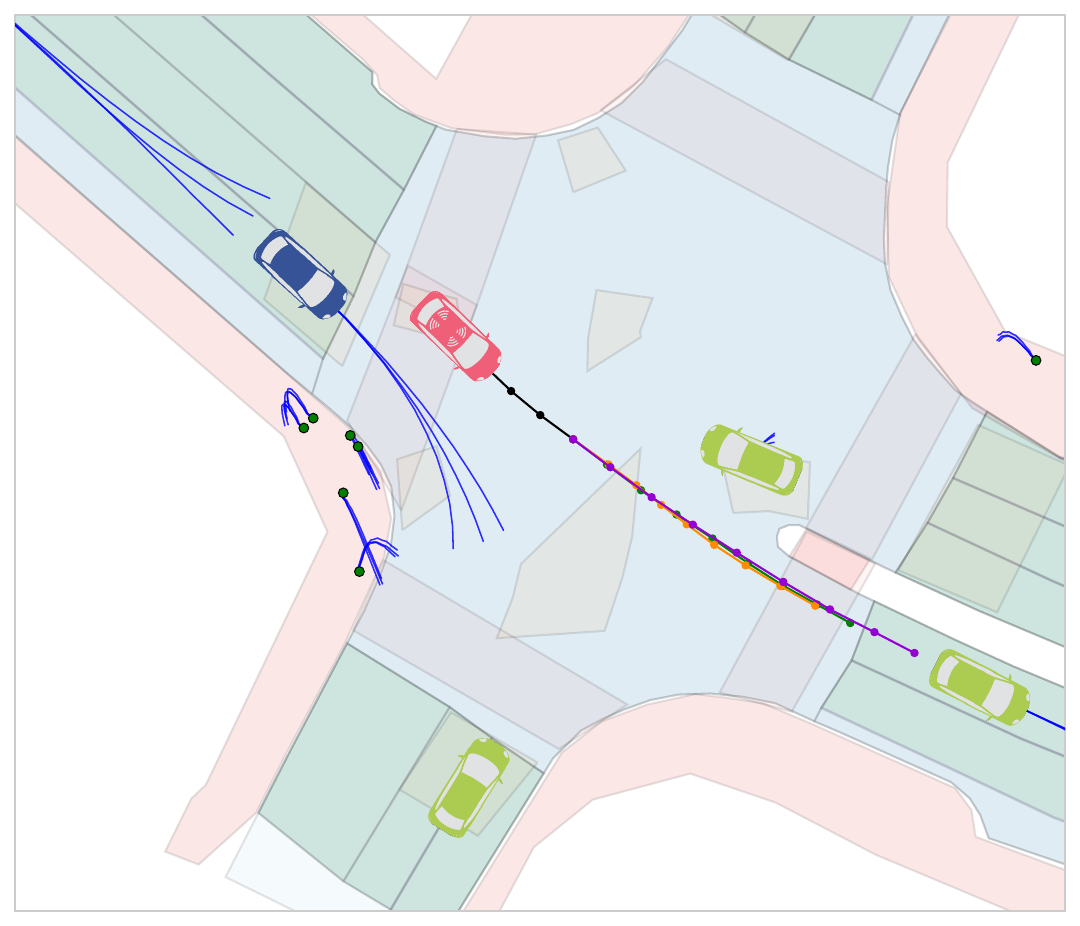}
    \includegraphics[width=0.29\linewidth]{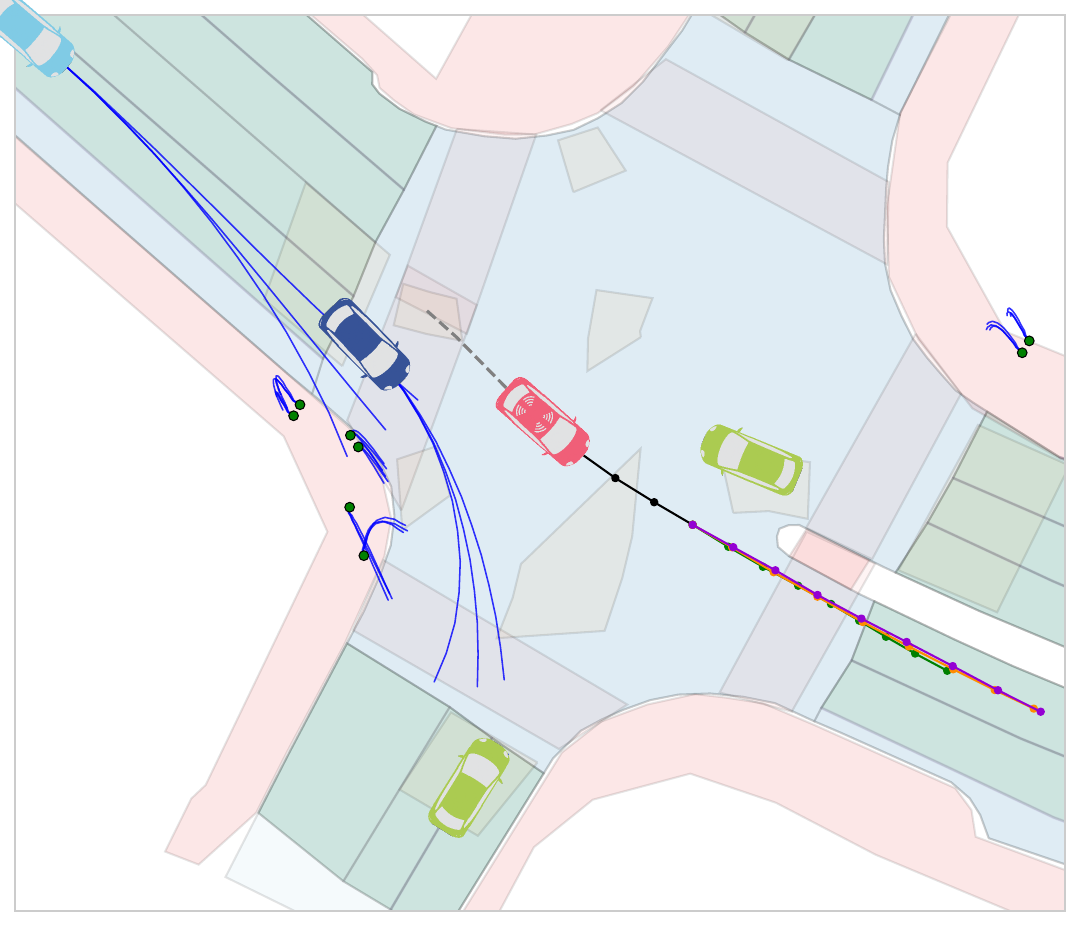}
    \includegraphics[width=0.29\linewidth]{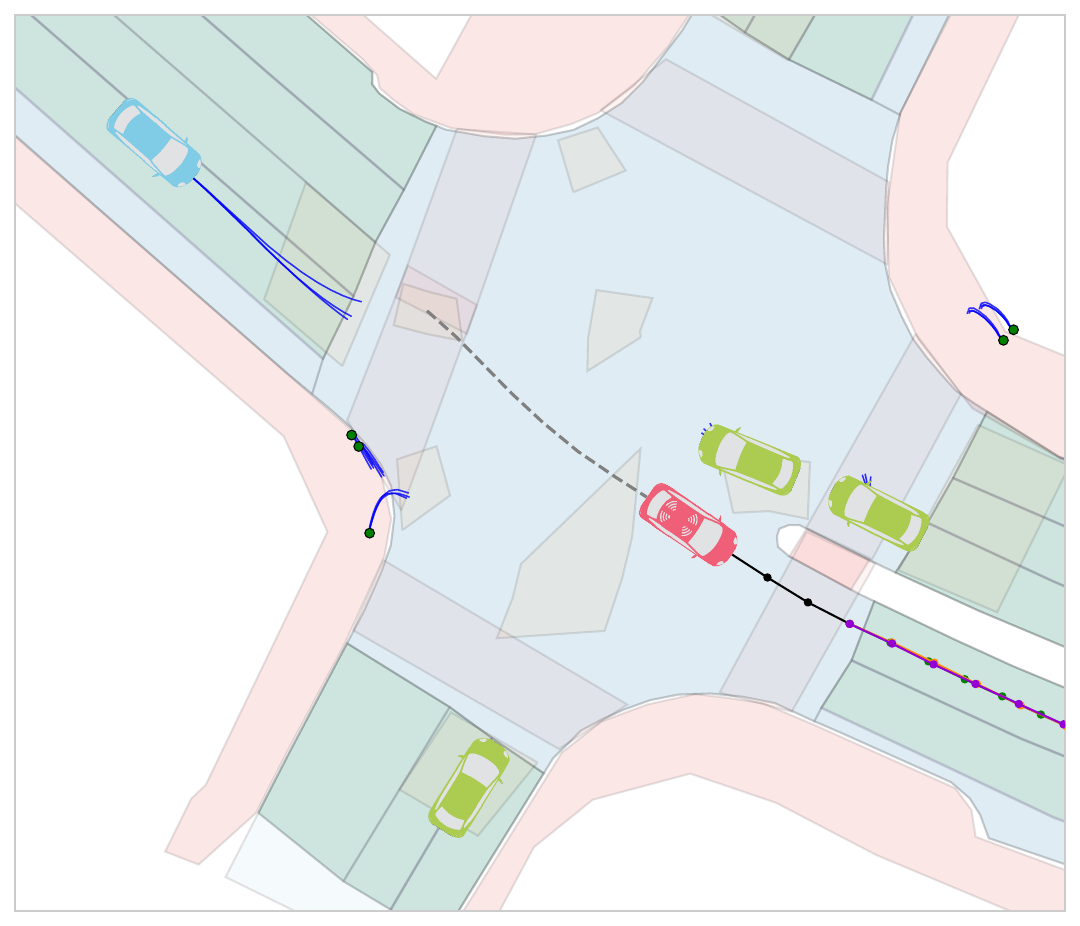}
    \raisebox{0.15\height}{\includegraphics[width=0.10\linewidth]{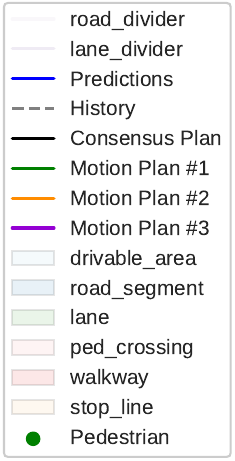}}
    \caption{Our combined prediction-planning method planning through an intersection. From left to right, we depict the scene at t = 0.25s, 1s, and 2s, respectively.}
    \label{fig:nuscenes_planning}
\end{figure}

In this section, we demonstrate that our combined predictor-planner framework both works in real-world scenarios and is more efficient than a state-of-the-art tracklet-based motion planner~\cite{SchmerlingLeungEtAl2018}. We instantiate the framework presented in \cref{sec:planning} by implementing an MPC planner adapted from~\cite{LamManzieEtAl2010,LinigerDomahidiEtAl2015}. The planner is tasked with guiding the ego-robot to follow its lane across an intersection while avoiding collisions with other agents. We model the ego-vehicle as a dynamically-extended unicycle (which was linearized) and plan using three parallel horizons (corresponding to the three most probable predicted modes) of horizon length 6.25s with a consensus horizon of 1s. A full description of the MPC formulation and implementation details can be found in the appendix. The resulting MPC problem was formulated as a convex Quadratic Program (QP) and solved using OSQP~\cite{StellatoBanjacEtAl2017}. Planning was performed in Julia with QP solve times averaging 50ms and PyTorch trajectory forecasting averaging 200ms on a 2.9 GHz dual-core Intel i5 CPU. 
The planner's behavior is shown in \cref{fig:nuscenes_planning} on an intersection crossing scenario from the nuScenes dataset. The ego-vehicle safely and smoothly crosses the intersection without excessively reacting to nearby agents or impeding traffic. The three planned horizons, which can be seen most clearly in the left two images differ primarily in longitudinal velocity. This is caused by differences in the predicted velocities of the vehicle leading it in its lane, visible in the left-most image.

\textbf{Comparison to Tracklet-Based Prediction and Planning.}
We compare our prediction and planning method to the recent tracklet-based prediction and planning method of~\cite{SchmerlingLeungEtAl2018}, which we refer to as TBPP. 
It plans maneuvers using a set of pre-determined robot action sequences and repeatedly queries its trajectory predictor to predict human responses. For the lane change scenario considered in~\cite{SchmerlingLeungEtAl2018}, TBPP takes 0.25s to predict and produce a single plan taking into account 1 other agent. In that same time budget, our method is able to predict and produce a plan for the scenario depicted in \cref{fig:nuscenes_planning}, accounting for 24 interacting agents as well as their 552 directed interactions! Notably, TBPP needs to generate nearly 100,000 future agent tracklets in parallel on a GPU through two rounds of model inference and beam search to obtain a sufficient covering of possible future outcomes per planning step. In comparison, our planner only queries the prediction model on a CPU once per planning step.

\section{Conclusion}
\label{sec:conclusion}

In this work, we present a novel multimodal output representation for trajectory forecasting, \algname{}, and propose a planning methodology that can take advantage of it to efficiently produce interaction-aware motion plans. In contrast to agent tracklet predictions which predict static trajectory distributions, a key benefit of \algname{} is that it is dynamic. Namely, a planner can explore the effect of any ego-robot control sequence post-hoc on the predicted agent behaviors in a scene. Its efficacy in prediction and planning has been demonstrated on the large-scale, real-world nuScenes dataset, yielding plans much more efficiently than a recent tracklet-based method. Future work includes incorporating scene context into the trajectory forecasting model to generate environment-aware \algname{} and improve prediction performance, as well as incorporating more sophisticated probabilistic reasoning and safety considerations into the planner. Finally, we plan to experimentally verify the performance of our combined prediction-planning framework on real-world hardware.

\clearpage
\acknowledgments{This work was supported by a NASA Space Technology Research Fellowship and the Toyota Research Institute (``TRI"). NASA and TRI provided funds to assist the authors with their research but this article solely reflects the opinions and conclusions of its authors and not NASA, TRI or any other Toyota entity.}

\bibliography{ASL_papers,main}  %

\clearpage
\newpage

\appendix

\section{Additional Training Information}

The Adam optimizer was used for both datasets, starting from an initial learning rate (specified below) which was then exponentially decayed every optimizer step with $\gamma = 0.9999$ to a minimum learning rate of $10^{-5}$. Further, we follow \cite{SalzmannIvanovicEtAl2020} and anneal $\beta$ in \cref{eqn:loss_fn} according to an increasing sigmoid. Initially, a low $\beta$ value is used so that the model learns to encode as much information in $z$ as possible early on. As training continues, $\beta$ is gradually increased to shift the role of information encoding from $q_\phi(z \mid \mathbf{x}, \mathbf{y})$ to $p_\theta(z \mid \mathbf{x})$. For $\alpha$, we found $\alpha = 1.0$ works well.

\subsection{Two Particle System}
The model was trained for 100 epochs, with an initial learning rate of $0.001$.

\subsection{nuScenes Dataset}
The model was trained for 10 epochs, with an initial learning rate of $0.002$. The model was trained to predict the next 6 timesteps (3s) having observed the previous 8 timesteps (4s). We used the nuScenes prediction challenge data splits and trained on the \texttt{train} set, tuning hyperparameters on the \texttt{train\_val} set, and evaluating on the \texttt{val} set.

To avoid overfitting to environment-specific characteristics, such as the general directions that agents move, we augment the data from each scene similar to \cite{SalzmannIvanovicEtAl2020}. Specifically, all trajectories in a scene are rotated around the origin from $0\degree$ to $345\degree$ (inclusive) in $15\degree$ intervals.

\section{Motion Planner Implementation Details}
The motion planner implementation is adapted from the MPCC controller in~\cite{LinigerDomahidiEtAl2015} to fit \cref{eqn:consensus_mpc} and results in a Quadratic Program (QP) to be solved at each sampling time. The planner is tasked with planning ego-vehicle trajectories that track a 2D reference path while avoiding agent positions predicted by the three most probable \algname{} modes. In this section, we attempt to follow the notational conventions of~\cite{LinigerDomahidiEtAl2015} where possible.

\textbf{Reference Path.} The reference path is given in terms of spatial coordinates with respect to some global frame and is parameterized by its arc length $\theta = [0,L]$, with $L$ representing the overall length. The coordinates of a reference point at some $\theta$ are denoted $X^\textrm{ref}(\theta)$ and $Y^\textrm{ref}(\theta)$. The path heading at a point (i.e. the angle of the tangent line with respect to the $X$-axis) is given by $\Phi(\theta)$. A reference path is produced by fitting a third order spline to the ego-vehicle's lane centerline.

\textbf{Tracking Error.} Let $[X, Y]^T$ represent the ego-vehicle's instantaneous position expressed in the global frame. We define $\theta_P$ as the arc length parameter yielding the reference point closest to the ego-vehicle, $[X^\textrm{ref}(\theta_P), Y^\textrm{ref}(\theta_P)]^T$. The vehicle's contouring (orthogonal) and lag (longitudinal) errors with respect to the reference path are respectively approximated by
\begin{align*}
    \hat{e}^c(X, Y, \theta_P) = \sin(\Phi(\theta_P))(X-X^\textrm{ref}(\theta_P)) - \cos(\Phi(\theta_P))(Y-Y^\textrm{ref}(\theta_P))\\
    \hat{e}^l(X, Y, \theta_P) = -\cos(\Phi(\theta_P))(X-X^\textrm{ref}(\theta_P)) - \sin(\Phi(\theta_P))(Y-Y^\textrm{ref}(\theta_P)).
\end{align*}
In order to penalize tracking error, we include a cost term in the planning objective approximating $q_c(\hat{e}^c(X, Y, \theta_P))^2 + q_l(\hat{e}^l(X, Y, \theta_P))^2$, where $q_c$ and $q_l$ are weighting parameters. As the contouring and lag error functions are nonlinear and would not be compatible with a QP formulation. Instead, we formulate an approximate tracking penalty by linearizing (i.e. performing a first-order Taylor expansion) $\hat{e}^{c,(t)}$ and $\hat{e}^{l,(t)}$ at each sampling time $t$ about some nominal point $[X_\textrm{nom}^{(t)}, Y_\textrm{nom}^{(t)}, \theta_\textrm{nom}^{(t)}]^T$. For the first iteration, these nominal points are chosen as evenly spaced points along the path corresponding to motion at some constant velocity. Subsequent iterations use the previous iteration's solution as the set of nominal points.  These linearized errors are then substituted into the original tracking penalty term (rather than their original, nonlinear forms) yielding a quadratic objective term of the form 
\begin{equation*}
\begin{bmatrix}
X^{(t)}\\
Y^{(t)}\\
\theta^{(t)}
\end{bmatrix}^T
\Gamma^{(t)}
\begin{bmatrix}
X^{(t)}\\
Y^{(t)}\\
\theta^{(t)}
\end{bmatrix}
+
\mathbf{l}^{(t)}
\begin{bmatrix}
X^{(t)}\\
Y^{(t)}\\
\theta^{(t)}
\end{bmatrix}
\end{equation*}
for each sampling time over the planning horizon.

\textbf{Ego-Vehicle Dynamics.} We model the ego-vehicle as a dynamically-extended unicycle \cite{LaValle2006BetterUnicycle} with dynamics given by
\begin{equation*}\label{eqn:supp_unicycle_continuous}
\begin{bmatrix}
\dot{X}\\
\dot{Y}\\
\dot{\phi}\\
\dot{v}
\end{bmatrix} = \begin{bmatrix}
v \cos (\phi)\\
v \sin (\phi)\\
\omega\\
a
\end{bmatrix}.
\end{equation*}
The ego-vehicle state consists of position coordinates $X$ and $Y$ (as defined previously), heading $\phi$ and longitudinal velocity $v$. The system is controlled by inputs $\omega$, the heading rate of change, and $a$, the longitudinal acceleration. Discrete-time dynamics are derived assuming a zero-order hold on the controls over each sampling interval (i.e. controls are piecewise constant) and are given by
\begin{equation*}\label{eqn:supp_unicycle}
\begin{aligned}
\begin{bmatrix}
X^{(t+1)}\\
Y^{(t+1)}\\
\phi^{(t+1)}\\
v^{(t+1)}
\end{bmatrix} &= \begin{bmatrix}
X^{(t)}\\
Y^{(t)}\\
\phi^{(t)}\\
v^{(t)}
\end{bmatrix} + \begin{bmatrix}
v^{(t)} \cdot D_S^{(t)} + \frac{a^{(t)} \sin(\phi^{(t)} + \omega^{(t)} \Delta t) \Delta t}{\omega^{(t)}} + \frac{a^{(t)}}{\omega^{(t)}} \cdot D_C^{(t)}\\
- v^{(t)} \cdot D_C^{(t)} - \frac{a^{(t)} \cos(\phi^{(t)} + \omega^{(t)} \Delta t) \Delta t }{\omega^{(t)}} + \frac{a^{(t)}}{\omega^{(t)}} \cdot D_S^{(t)}\\
\omega^{(t)} \Delta t\\
a^{(t)} \Delta t
\end{bmatrix},\\
\end{aligned}
\end{equation*}
where $\Delta t$ represents the length of the sampling interval and where
\begin{equation*}
D_S^{(t)} = \frac{\sin(\phi^{(t)} + \omega^{(t)} \Delta t) - \sin(\phi^{(t)})}{\omega^{(t)}} \text{ and }
D_C^{(t)} = \frac{\cos(\phi^{(t)} + \omega^{(t)} \Delta t) - \cos(\phi^{(t)})}{\omega^{(t)}}.
\end{equation*}
To incorporate dynamics constraints within the QP, the discrete-time dynamics are linearized at each sampling time about some nominal state and control vectors, $\mathbf{q}_\textrm{nom}^{(t)}$ and $\mathbf{u}_\textrm{nom}^{(t)}$, respectively. This results in constraints of the form 
\begin{equation}\label{eqn:linearized_veh_dynamics}
\mathbf{q}^{(t+1)} = A_\textrm{ego}^{(t)}\mathbf{q}^{(t)} + B_\textrm{ego}^{(t)}\mathbf{u}^{(t)} + \mathbf{c}_\textrm{ego}^{(t)}
\end{equation}
at the beginning of each sampling interval over the planning horizon. Note that $A_\textrm{ego}^{(t)}$, $B_\textrm{ego}^{(t)}$ and $\mathbf{c}_\textrm{ego}^{(t)}$ are the results of the linearization about $\mathbf{q}_\textrm{nom}^{(t)}$ and $\mathbf{u}_\textrm{nom}^{(t)}$.

To link the ego-vehicle state to the reference path, and thus be able to compute a tracking error, we introduce the so-called "path dynamics,"
\begin{equation} \label{eqn:path_dynamics}
    \theta^{(t+1)} = v_s^{(t)}\Delta t + \theta^{(t)}.
\end{equation}
The path dynamics employ a first-order integration scheme to approximate the ego-vehicle's arc length parameter at each sampling time, $\theta^{(t)}$, over the planning horizon. We use $v_s$ to represent the velocity projected along the reference path. For the purposes of planning, we consider $\theta$ to be an additional ego-vehicle state component and $v_s$ to be an additional control input.

\textbf{Collision Avoidance.}
For a particular mode $z \in Z$, the dynamics of all agents in a scene are predicted by \cref{eqn:uncertain_ATV}, permitting the prediction of agent positions at different sampling times. We use $X_j^{(t)}$ and $Y_j^{(t)}$ to represent the mean predicted position of agent $j \in J$ at time $t$, where $J$ represents the set of agents in the scene. Let $n_j{(t)}^{(t)}$ represent the normal vector from the position of agent $j$ to the ego-vehicle position at time $t$. Obstacle avoidance constraints are enforced as half-plane constraints of the form 
\begin{equation*}
n_j^{(t)} 
\begin{bmatrix}
X^{(t)}\\
Y^{(t)}\\
\end{bmatrix} \geq d,
\end{equation*} 
for all $j \in J$ and for all sampling times $t$ over the planning horizon, with $d$ representing a distance margin.


\textbf{Planning Problem.} We seek motion plans that guide the ego-vehicle to follow its current lane, while avoiding other agents. We consider predictions from the three most probable modes $Z = \{z_1, z_2, z_3\}$. The MPC optimization problem is formulated as

\begin{align*}
\begin{split}
\min_{\mathbf{\bar{s}},\mathbf{\bar{u}}_{\text{R}}} \quad &\sum_{z \in Z}\begin{bmatrix}
X_{z}^{(T)}\\
Y_{z}^{(T)}\\
\theta_{z}^{(T)}
\end{bmatrix}^T
\Gamma_{z}^{(T)}
\begin{bmatrix}
X_{z}^{(T)}\\
Y_{z}^{(T)}\\
\theta_{z}^{(T)}
\end{bmatrix}
+
\mathbf{l}_{z}^{(T)}
\begin{bmatrix}
X_{z}^{(T)}\\
Y_{z}^{(T)}\\
\theta_{z}^{(T)}
\end{bmatrix} + \sum_{z \in Z}\sum_{t=1}^{T-1}\begin{bmatrix}
X_{z}^{(t)}\\
Y_{z}^{(t)}\\
\theta_{z}^{(t)}
\end{bmatrix}^T
\Gamma_{z}^{(t)}
\begin{bmatrix}
X_{z}^{(t)}\\
Y_{z}^{(t)}\\
\theta_{z}^{(t)}
\end{bmatrix}
+
\mathbf{l}_{z}^{(t)}
\begin{bmatrix}
X_{z}^{(t)}\\
Y_{z}^{(t)}\\
\theta_{z}^{(t)}
\end{bmatrix} +\\ &q_u \Delta \mathbf{u}_z^{(t)} - \gamma v_{s,z}^{(t)} \\
\\
\textrm{s.t.} \quad &\mathbf{s}_z^{(t + 1)} = A_z^{(t)} \mathbf{s}_z^{(t)} + B_z^{(t)} \mathbf{u}^{(t)}_{\text{R}} + \mathbf{c}_z^{(t)} \ \forall\ i \in \{1,2,3\}\\ 
&\mathbf{s}_z^{(1)} = \mathbf{s_{\textrm{init}}},\ \mathbf{s}_z^{(t)} \in \mathcal{S}^{(t)},\  \mathbf{u}_{\text{R}}^{(t)} \in \mathcal{U}^{(t)},\ \forall\ z \in Z\\ 
&n_{z,j}^{(t)} 
\begin{bmatrix}
X_z^{(t)}\\
Y_z^{(t)}\\
\end{bmatrix} \geq d,\ \forall\ j \in J, z \in Z\\
&\mathbf{u}_{\text{R}, z_1}^{(t)} = \mathbf{u}_{\text{R}, z_2}^{(t)}\ \forall\ z_1, z_2 \in Z,\ t \in \{1,2,\ldots,t_c\}.
\end{split}
\end{align*}
As before, the decision variables for this problem are $\mathbf{\bar{s}} = \bigcup_{z \in Z} (\mathbf{s}_z^{(1)}, \ldots, \mathbf{s}_z^{(T)})$ and $\mathbf{\bar{u}_R} = \bigcup_{z \in Z} (\mathbf{u}_{\text{R},z}^{(1)}, \ldots, \mathbf{u}_{\text{R},z}^{(T-1)})$, representing the state and control trajectories across all modes indexed by $z \in Z$. A slew-rate penalty term, $\Delta \mathbf{u}_z^{(t)}$, weighted by $q_u$, is included in the objective function to encourage smooth controls. Additionally, we introduce $\gamma v_{s,z}^{(t)}$ as a reward for progress along the path. Note that \cref{eqn:linearized_veh_dynamics} and \cref{eqn:path_dynamics} are assumed to constitute the "overall" ego-vehicle dynamics and are included within the first constraint. 

\textbf{Planning Details.} Planning is performed with a horizon of 3s with a $\Delta t$ of 0.25s (i.e. 12 steps). A consensus horizon $t_c$ of 1s was enforced (i.e. 4 steps). The admissible controls are restricted to $-0.7\ \text{rad/s} \leq \omega \leq 0.7\ \text{rad/s}$ and $-5\ \text{m/s}^2 \leq a \leq 4\ \text{m/s}^2$. Velocity was restricted to $0.05\ \text{m/s} \leq v \leq 12\ \text{m/s}$ as a state constraint. For penalty weights, we use $q_c = 0.5$, $q_l = 0.5$, $q_u = 0.01$ and $\gamma = 0.02$.

\section{Additional Frobenius Norm Visualizations}

\begin{figure}[t]
    \centering
    \includegraphics[width=0.36\linewidth]{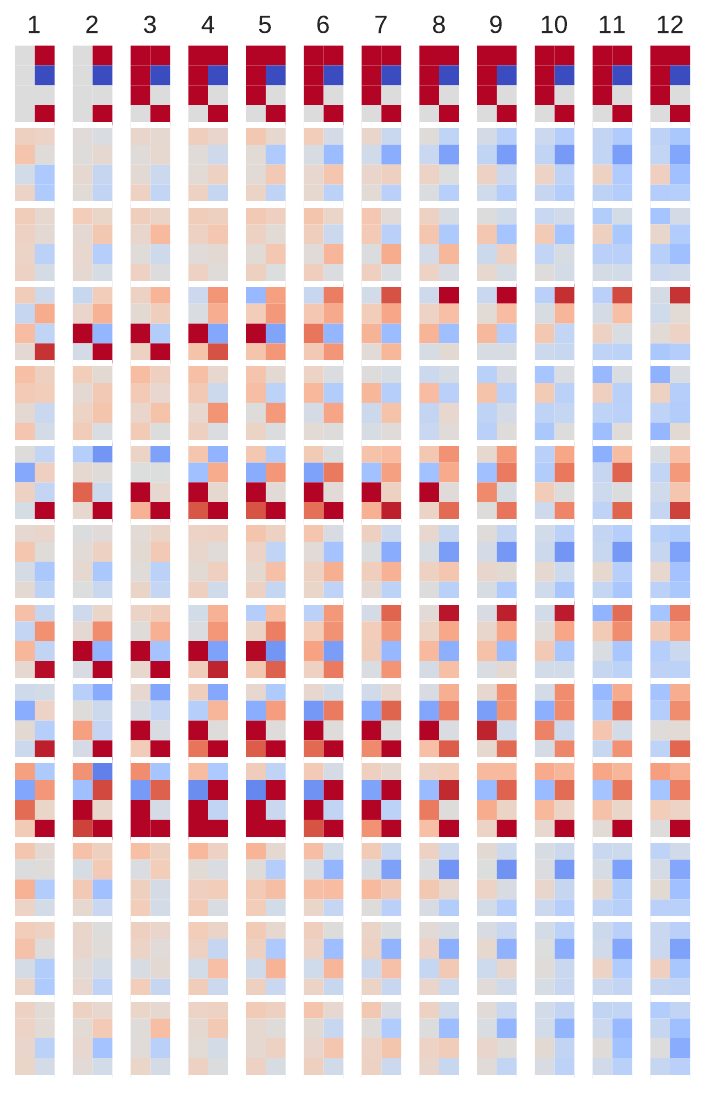}
    \includegraphics[width=0.63\linewidth]{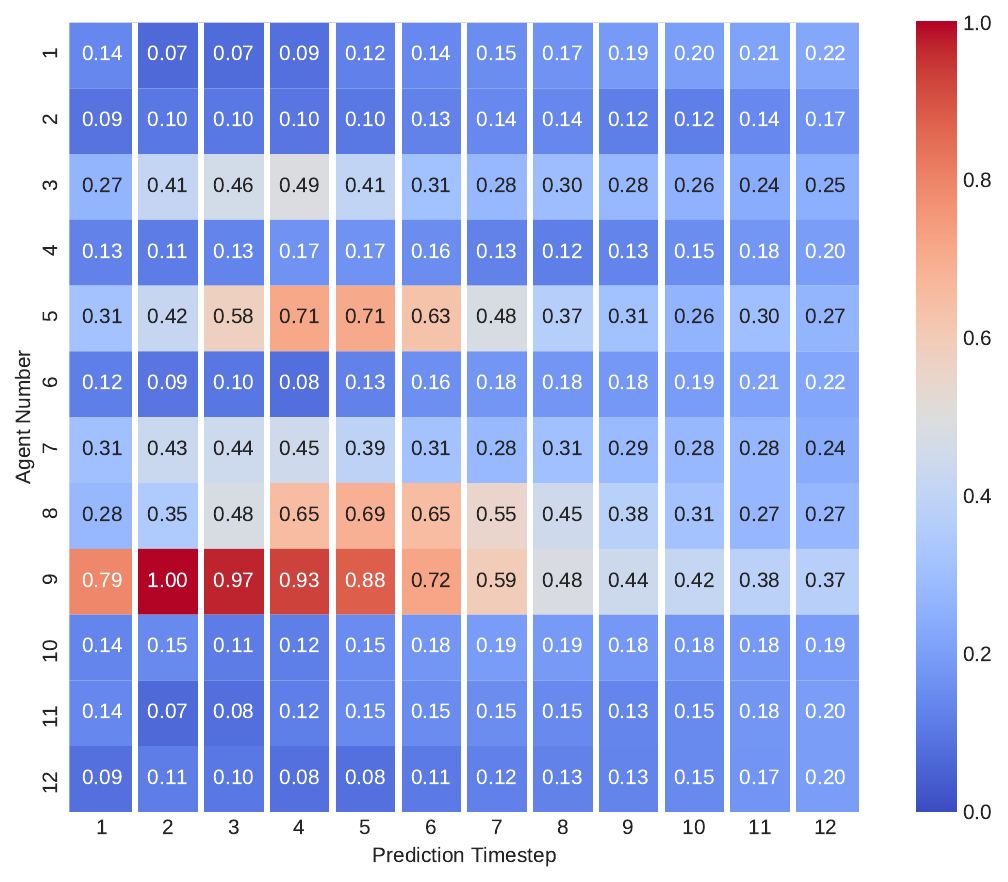}
    \caption{\textbf{Left:} The set of $B^{(t)}_z$ matrices across time from the most likely $z$ mode for the intersection scene shown in \cref{fig:nuscenes_B_graph}. White horizontal lines separate each agent's block, e.g., the first $4 \times 2$ block is the ego-robot's block and the next $4 \times 2$ block belongs to an agent being modeled. \textbf{Right:} The Frobenius norm of each non-robot agent's block in the $B^{(t)}_z$ matrix is visualized per prediction timestep, normalized so the largest Frobenius norm is $1$.}
    \label{fig:supp_B_mags}
\end{figure}

\cref{fig:supp_B_mags} shows the set of $B^{(t)}_z$ matrices from the most likely $z$ mode for the scene depicted in \cref{fig:nuscenes_B_graph}, where clear temporal patterns can be seen in five of the submatrices. These five correspond to the five other vehicles in the scene, and this pattern is caused by the ego-vehicle crossing through an intersection, which is a time of maximal influence on the other vehicular agents. Specifically, the agents numbered 3, 5, 7, 8, 9 in the right of \cref{fig:supp_B_mags} are vehicles and the rest are pedestrians. Notably, the pedestrian agents interact weakly with the ego-vehicle, indicating that they are acting independently from the vehicle's motion. This makes sense as the pedestrians in \cref{fig:nuscenes_B_graph} are walking on the sidewalk or standing still waiting at a crosswalk, unaffected by the ego-vehicle.

\end{document}